\documentclass[runningheads]{llncs}
\usepackage{amsmath,amssymb,amsfonts}
\usepackage{graphicx}
\usepackage{float}
\usepackage{multirow}
\usepackage{textcomp}
\usepackage{pifont}
\usepackage{algorithmic}
\usepackage[export]{adjustbox}
\begin{document}
\title{GMA3D: Local-Global Attention Learning to Estimate Occluded Motions of Scene Flow}
%
%
\author{Zhiyang Lu \and
Ming Cheng }
\institute{Xiamen University}
%


\maketitle              
\begin{abstract}
Scene flow represents the motion information of each point in the 3D point clouds. It is a vital downstream method applied to many tasks, such as motion segmentation and object tracking. However, there are always occlusion points between two consecutive point clouds, whether from the sparsity data sampling or real-world occlusion. In this paper, we focus on addressing occlusion issues in scene flow by the semantic self-similarity and motion consistency of the moving objects. We propose a GMA3D module based on the transformer framework, which utilizes local and global semantic similarity to infer the motion information of occluded points from the motion information of local and global non-occluded points respectively, and then uses an offset aggregator to aggregate them. Our module is the first to apply the transformer-based architecture to gauge the scene flow occlusion problem on point clouds. Experiments show that our GMA3D can solve the occlusion problem in the scene flow, especially in the real scene. We evaluated the proposed method on the occluded version of point cloud datasets and get state-of-the-art results on the real scene KITTI dataset. To testify that GMA3D is still beneficial to non-occluded scene flow, we also conducted experiments on non-occluded version datasets and achieved promising performance on FlyThings3D and KITTI. The code is available at https://anonymous.4open.science/r/GMA3D-E100.

\end{abstract}

\section{Introduction}

It is significant to capture object motion information in dynamic scenes. Scene flow~\cite{vedula1999three} calculates the motion field of two consecutive frames of the 3D scenes and obtains collections of directions and distances of object movements. Scene flow is the underlying motion information, which is serviceable in many applications, such as robotic path planning, object tracking, and augmented reality. The previous methods use RGB images~\cite{huguet2007variational,pons2007multi,wedel2008efficient,wedel2011stereoscopic,basha2013multi,vcech2011scene,vogel20113d,vogel2013piecewise,vogel20153d} to estimate the scene flow, but 2D methods cannot accurately consider the 3D information in the real scenes. Instead, with the advances in 3D sensors, it is easy to obtain point cloud data. PointNet~\cite{qi2017pointnet} and PointNet++~\cite{qi2017pointnet++} pioneered the direct extraction of features for raw point clouds, and then the deep learning networks~\cite{guo2021pct-tf-pc,li2018pointcnn,qi2017pointnet,qi2017pointnet++,wang2019dynamic,wu2019pointconv,shi2020pv-rcnn} in the field of point clouds continued to emerge. These works provide the necessary conditions for the scene flow task. Combined with these frameworks, many neural network architectures suitable for scene flow~\cite{liu2019flownet3d,gu2019hplflownet,wu2020pointpwc,wei2021pv,kittenplon2021flowstep3d} are proposed, which have better performance than the traditional optimization-based methods. Although these methods have gained good results on the non-occluded datasets, they fail to infer the motion information of the occluded objects, which will lead to the scene flow deviation in large-scale occlusion scenes, such as in large-scale traffic jams.

In the scene flow task, the occluded points exist in the first frame (source) point cloud. We define it as a set of points without corresponding points and/or corresponding patches in the second frame (target). Furthermore, we divide occluded points into two categories: the first category has non-occluded points in local areas of the first frame point cloud, and these points are called local occluded points. The second kind of point is global occlusion points, where there are no non-occluded points in their local areas. The previous method calculates the corresponding scene flow through the feature matching between two frames, which can well infer the scene flow of non-occluded points in the first frame, because such points have corresponding matching patches in the second frame, and the motion information can be deduced through the cross-correlation between the point clouds of the two frames. However, the occluded points have no corresponding matching patches in the second point cloud, so it is incapable to infer the motion information by cross-correlation. In contrast, humans often employ self-correlation when deducing the motion of occluded objects in dynamic scenes. For example, without considering collision, we can infer the motion information of the occluded head of the same vehicle from the tail. Therefore, the self-correlation of motion is very significant to solve the occlusion problem in scene flow.

Previously, Ouyang et al. combined scene flow estimation task with occlusion detection task~\cite{ouyang2021ogsf}, and optimized the two target tasks to infer the motion information of occluded points. Such a method can effectively cure local small-scale occlusion issues, but it still cannot resolve the problem of local large-scale occlusion and global occlusion. Jiang et al.~\cite{jiang2021learning-gma} designed a transformer-based global motion aggregation (GMA) module to conclude the motion information of occluded pixels in optical flow. Inspired by this, we propose GMA3D, which integrates transformer~\cite{vaswani2017attentionisallyouneed} framework into scene flow tasks, utilizing the self-similarity of point clouds features to aggregate motion features and obtain the motion information of occluded points. Unfortunately, previous works only consider motion features from the global perspective without regarding the local consistency of motion, which may lead to error motion of local occlusion points.

\begin{figure}
\includegraphics[width=\textwidth]{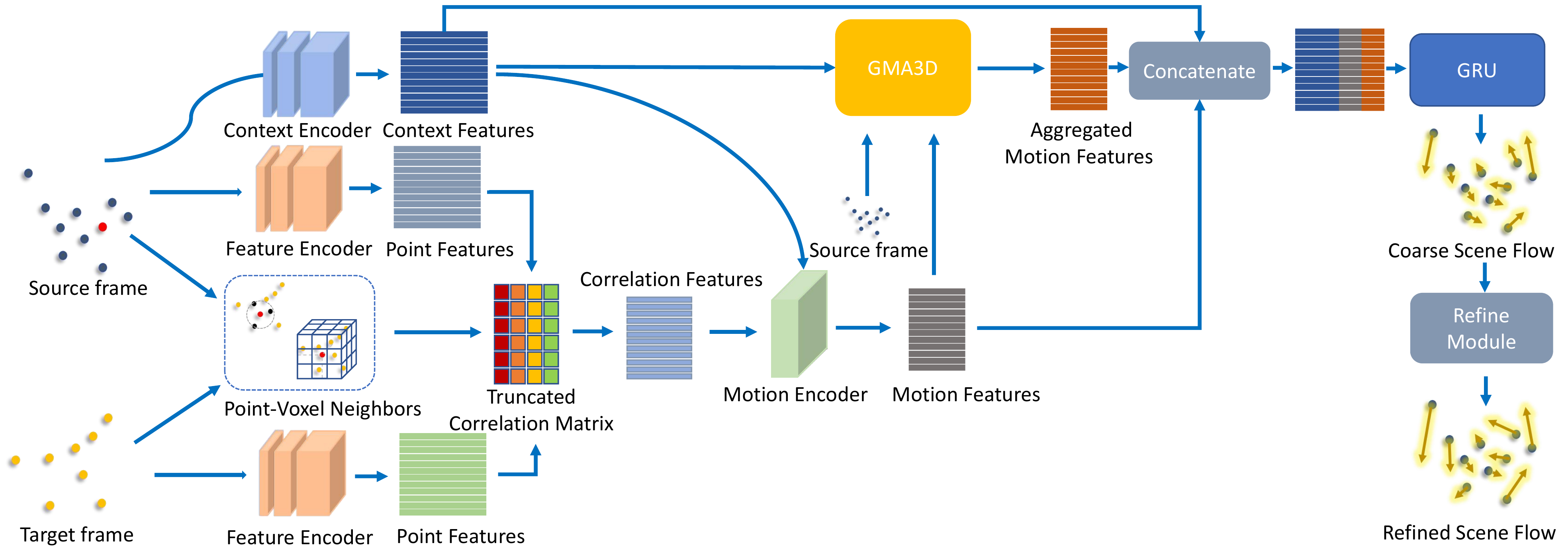} 
\caption{The overall pipeline of our proposed framework. Our network is based on the successful PV-RAFT~\cite{wei2021pv} architecture. The input of the GMA3D module is the context features and motion features of the point cloud in the first frame, and the output is the motion features aggregated locally and globally. These aggregated motion features are concatenated with context features and the original motion features, and then the concatenated features are fed into GRU for residual flow estimation, which is finally refined by the refine module.} \label{fig1}
\end{figure}

To address these issues, we present a local-global semantic similarity map (LGSM) module to calculate the local-global semantic similarity map and then employ an offset aggregator (OA) to aggregate motion information based on self-similarity matrices. For local occlusion, we deduce the motion information of the occluded points from their local non-occluded neighbors based on local motion consistency. As far as global occlusion points, we apply the global semantic features to aggregate motion features from non-occluded points. We utilize these local and global aggregated motion features to augment the successful PV-RAFT~\cite{wei2021pv} framework and achieve state-of-the-art results in occluded scene flow estimation.

The key contributions of our paper are as follows. We propose a transformer-based framework GMA3D to address the problem of motion occlusion in scene flow, in which we designed the LGSM module to leverage the self-consistency of motion information from both local and global perspectives, and then apply the offset aggregator to aggregate the motion features of the non-occluded points with self-similarity to the occluded points. Moreover, we demonstrate that the GMA3D module reduces the local motion bias by aggregating local and global motion features, which is also beneficial to non-occluded points. Experiments have shown that our GMA3D module has attained exceptional results in scene flow tasks, whether in the case of occluded or non-occluded datasets.

\section{Related Work}

\subsection{Motion Occlusion of Scene Flow }
There are few techniques to address the occlusion problem of scene flow. Self-Mono-SF~\cite{Hur_2020_self_mo_sf} utilizes self-supervised learning with 3D loss function and occlusion reasoning to infer the motion information of occlusion points in monocular scene flow. \cite{ilg2018occlusions-motion-depth-boundary} combines occlusion detection, depth, and motion boundary estimation to infer occlusion points and scene flow. PWOC-3D~\cite{saxena2019pwoc} constructs a compact CNN architecture to predict scene flow in stereo image sequences and proposes a self-supervised strategy to produce the occlusion map for improving the accuracy of flow estimation. OGSF~\cite{ouyang2021ogsf} presents the same backbone network to optimize scene flow and occlusion detection, then integrate the occlusion detection results with the cost volume between two frame point clouds, changing the cost volume of occlusion points to 0. However, it relies on occlusion mask ground truth in the scene flow dataset. This paper will propose a module designed to estimate motion information of occluded points, which can be seamlessly integrated into any scene flow network architecture.

\section{Methodology}

\subsection{Problem Statement}
We consider scene flow as a 3D motion estimation task. It inputs two consecutive frames of point cloud data $P C_{t}=\{p c_{t}^{i} \in \mathbb{R}^{3}\}_{i=1}^{N}$ and $P C_{t+1}=\{p c_{t+1}^{j} \in \mathbb{R}^{3}\}_{j=1}^{M}$ and outputs the 3D vector $Flow=\{f_{i} \in \mathbb{R}^{3}\}_{i=1}^{N}$ of each point in the first frame of $P C_{t}$ to indicate how to move to the corresponding position of the second frame.

\subsection{Background}
The backbone architecture of our GMA3D module is PV-RAFT~\cite{wei2021pv}. The overall network diagram is shown in Figure~\ref{fig1}. For completeness, we will briefly introduce the PV-RAFT model. PV-RAFT adopts the point-voxel strategy to calculate the cost volume of the source point cloud. At the point level, the KNN method is used to find the points in the neighborhood of the target point cloud for short-distance displacement. At the voxel level, the points in the target point cloud are voxelized based on the source point cloud to capture the long-distance displacement. Then, it sends the point cloud context features together with the cost volumes into the GRU-based iteration module to estimate the residual flow. Finally, the flow features are smoothed in refine module. However, PV-RAFT removes the occlusion points when processing datasets, so it is unable to address the occlusion problem in scene flow.

\begin{figure}
\includegraphics[width=\textwidth]{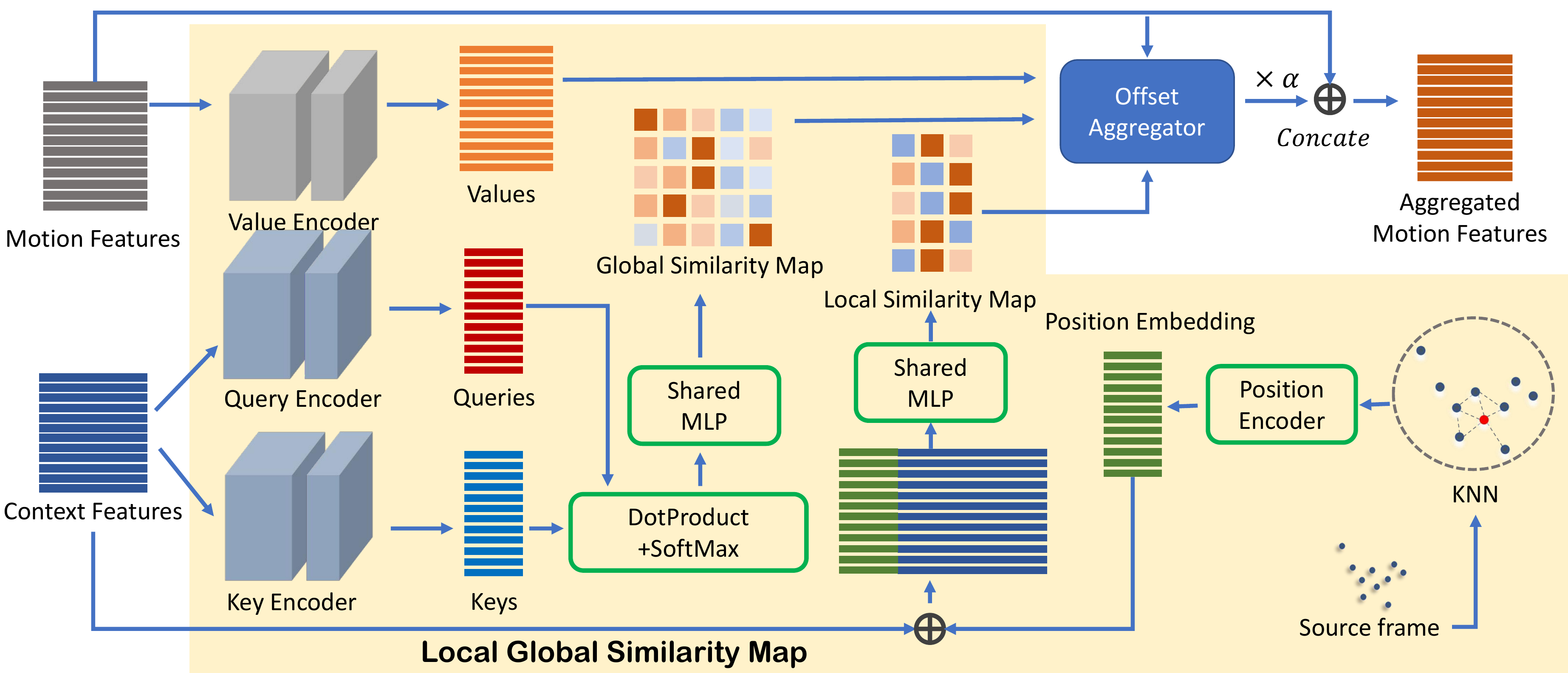} 
\caption{Illustration of GMA3D module details. We use the LGSM module to map the local and global semantic similarity of the point cloud in the first frame. In the LGSM module, we map the context features to the query feature map and key feature map by a linear model with shared weights. Next, the attention map produced by the dot product is applied with softmax and Muti-Layer Perception(MLP) to generate the global self-similarity map. Then, the local similarity map is calculated by utilizing a Local GNN to model the positional relations in Euclidean space among the first point clouds. Finally, the local and global semantic similarity map are weighted sum with the motion features projected from the value encoder through the offset aggregator to output local and global aggregated motion features.} 
\label{fig2} 
\end{figure}

\subsection{Overview}
In the optical flow task, GMA~\cite{jiang2021learning-gma} leverages the transformer to globally aggregate the motion features of similar pixels to infer the motion information of the occluded pixels. Inspired by GMA, we utilize the self-similarity method to solve the occlusion problem of scene flow, which has never been used in the scene flow area. We map the context features to query and key features through the linear model with shared weights, map the motion features to value features through another linear model, and then utilize a transformer-based framework to aggregate the motion features.

However, GMA is only dependent on global feature similarity to aggregate motion features, which may lead to some motion deviations. For example, there are many vehicles with similar features in a street scene, but their motions may be diverse. If only global feature similarity is used for motion information aggregation, the motion of a reverse-moving vehicle may be incorrectly aggregated to the occlusion position of another vehicle. Therefore, when solving the problem of motion occlusion, we also need to consider the consistency of local motion. The closer the relative distance between points in the same frame point cloud with similar contextual features, the more consistent the motion information is. We integrate a Local GNN into the similarity map and propose a local-global semantic similarity map module, which is used to aggregate local and global motion features respectively. The aggregated local-global motion features are concatenated with the original motion features and the context features and then fed into the GRU module to iteratively estimate the scene flow. The detailed diagram of our GMA3D module is demonstrated in Figure~\ref{fig2}.

\subsection{Mathematical Formulation}
Let $\textit{q}, \textit{k}, \textit{v}$ be the \textit{query}, \textit{key} and \textit{value} projection operators respectively, the formula is as follows:
\begin{equation}
    q(\mathbf{x}_i) = \mathbf{Q}_{m}\mathbf{x}_{i} 
\end{equation}
\begin{equation}
   k(\mathbf{x}_j) = \mathbf{K}_{m}\mathbf{x}_{j}  
\end{equation}
\begin{equation}
    v(\mathbf{y}_j) = \mathbf{V}_{m}\mathbf{y}_{j}    
\end{equation}
In these formulas, 
$\textbf{x}=\{\mathbf{x}_{i}\}_{i=1}^{N}\in{\mathbb{R}^{N\times{D_c}}}$ 
denote the context features and $\textbf{y}=\{\mathbf{y}_{j}\}_{j=1}^{N}\in{\mathbb{R}^{N\times{D_m}}}$ 
refer to the motion features, where $N$ is the number of 
\begin{figure}
\includegraphics[width=\textwidth]{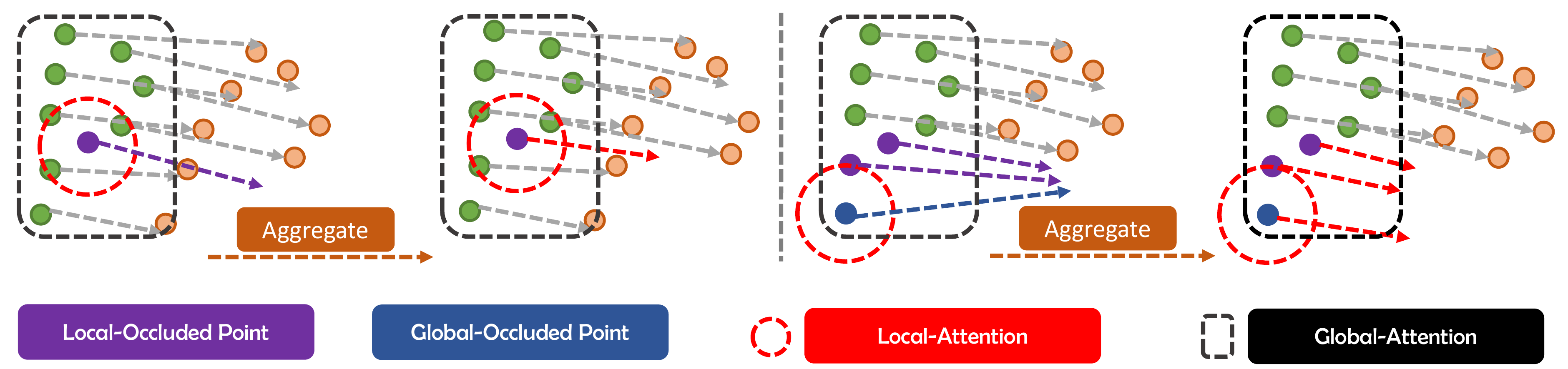} 
\caption{Local-Global attention mechanism for GMA3D in local occluded points (left) and global occluded points (right)} 
\label{local-global} 
\end{figure}
the source point cloud, $D_c$ and $D_m$ refer to the dimension of context features and motion features respectively. Moreover, $\mathbf{Q}_m,\mathbf{K}_m\in{\mathbb{R}^{D_c\times{D_{q,k}}}}$ is a shared learnable linear projection  and $\mathbf{V}_m\in{\mathbb{R}^{D_m\times{D_{m}}}}$.

First, we project the contextual information to the query map and key map employing $q(x)$ and $k(x)$, and compute the local similarity map by Local-GNN and global similarity map by the function $f(x,y)$ separately. Then, we map the motion features into value features by $v(y)$ and generate local and global aggregated motion features by local and global semantic similarity map, respectively.
\begin{equation}
\mathbf{g}_{local} = \sum_{j\in{N(x_i)}}
MLP([MLP(pc_{t+1}^{j}-pc_{t}^{i}) ,\mathbf{x}_j,\mathbf{x}_i]))v(\mathbf{y}_j)
\end{equation}
\begin{equation}
\mathbf{g}_{global} = \sum_{j\in{P_{source}}}f(q(\mathbf{x}_i),k(\mathbf{x}_j))v(\mathbf{y}_j)
\end{equation}
Here $N(x_i)$ is the set of local neighborhood points of $x_i$ captured by KNN,'[·,·]'denotes the concatenation operator and $f$ represents the operation given by
\begin{equation}
\bar{x}_{i,j}=\frac{\exp \left(\mathbf{x}_{i}^{\top} \mathbf{x}_{j}\right)}{\sum_{j=1}^{N} \exp \left(\mathbf{x}_{i}^{\top} \mathbf{x}_{j}\right)}
\end{equation}
\begin{equation}
f\left(\mathbf{x}_{i}, \mathbf{x}_{j}\right) = MLP(\frac{\bar{x}_{i, j}}{\sum_{k} \bar{x}_{i, k}})  
\end{equation}
Finally, we apply the offset aggregator to get the local and global aggregated motion information and add it to the original motion information according to the learnable coefficient to obtain the final output.
\begin{equation}
\mathbf{g}_{offset} = h_{l,b,r}(\mathbf{y}-(\mathbf{g}_{local}+\mathbf{g}_{global}))
\end{equation}
\begin{equation}
\tilde{y_i} = y_i + \alpha(\mathbf{g}_{offset})  
\end{equation}
where $h_{l,b,r}$ refers to linear model, batch-norm and relu. 

\begin{table}[hbt]
\begin{center}
\caption{ Performance comparison on the FlyingThings3Ds and KITTIs datasets. All the models in the table are only trained on the non-occluded Flyingthings3D in a supervised manner and tested on the non-occluded KITTI without fine-tune. The best results for each dataset are marked in bold.}
\label{table:s} 

\begin{tabular}{c|l|llll}
\hline
Dataset & Method & EPE(m)${\downarrow}$ & Acc Strict${\uparrow}$ & Acc Relax${\uparrow}$  & Outliers${\downarrow}$ \\
\hline

\multirow{5}{*}{FlyThings3Ds} & HPLFlowNet\cite{gu2019hplflownet} & 0.0804 & 0.6144 & 0.8555 & 0.4287 \\

& PointPWC-Net\cite{wu2020pointpwc} & 0.0588 & 0.7379 & 0.9276 & 0.3424 \\

& FLOT\cite{puy2020flot} & 0.0520 & 0.7320 & 0.9270 & 0.3570 \\

& PV-RAFT(baseline)\cite{wei2021pv} & 0.0461 & 0.8169 & 0.9574 & 0.2924 \\

& GMA3D & \textbf{0.0397} & \textbf{0.8799} & \textbf{0.9727} & \textbf{0.2293} \\
\hline

\multirow{5}{*}{KITTIs} 
& HPLFlowNet\cite{gu2019hplflownet} & 0.1169 & 0.4783 & 0.7776 & 0.4103 \\

& PointPWC-Net\cite{wu2020pointpwc} & 0.0694 & 0.7281 & 0.8884 & 0.2648 \\

& FLOT\cite{puy2020flot} & 0.0560 & 0.7550 & 0.9080 & 0.2420 \\

& PV-RAFT(baseline)\cite{wei2021pv} & 0.0560 & 0.8226 & 0.9372 & 0.2163 \\

& GMA3D & \textbf{0.0434} & \textbf{0.8653} & \textbf{0.9692} &\textbf{0.1769}  \\
\hline
\end{tabular}

\end{center}
\end{table}

\section{Experiments}

\subsection{Datasets}
 Following previous methods~\cite{wei2021pv,kittenplon2021flowstep3d,puy2020flot,ouyang2021ogsf}, we trained our model on the FlyThings3D~\cite{mayer2016large-fly3d} dataset and tested it on both FlyThings3D and KITTI~\cite{menze2015joint-kitti2015,menze2015object-kitti2015} datasets respectively. Currently, there are two different approaches to process these datasets, so we compare them separately on datasets generated by these different processing methods. The first method is derived from~\cite{gu2019hplflownet}, in which the occluding point and some difficult points are removed. Following~\cite{puy2020flot}, we call this version of these datasets FlyThings3Ds and KITTIs. Another way to obtain the scene flow datasets comes from~\cite{liu2019flownet3d}, where the information on occlusion points is preserved. We refer to the second version of these datasets as FlyThings3Do and KITTIo. But unlike the previous method~\cite{wei2021pv,liu2019flownet3d,kittenplon2021flowstep3d,puy2020flot}, we trained on all points including the occlusion points in the datasets of the occluded version to demonstrate that our GMA3D module can be used to solve the occlusion problem in the scene flow.

\subsection{Evaluation Metrics}
Adhere to the previous methods~\cite{liu2019flownet3d,gu2019hplflownet,puy2020flot,wei2021pv}, we still employed traditional evaluation operators to compare the performance of our GMA3D module, including EPE(m), Acc strict, Acc relax, and Outliers:
\begin{itemize}
    \item \textbf{EPE(m)}: $\|f_{p r e d}-f_{g t}\|_{2}$. Average of the end-point-error at each point.
    \item \textbf{Acc Strict}: the percentage of points whose \textbf{EPE(m)} $< 0.05m$ or relative error $< 5\%$.
    \item \textbf{Acc Relax}: the percentage of points whose \textbf{EPE(m)} $< 0.1m$ or relative error $< 10\%$.

    \item \textbf{Outliers}: the percentage of points whose \textbf{EPE(m)} $> 0.3m$ or relative error $> 30\%$.
\end{itemize}

\begin{figure}[tb]
\centering 
\includegraphics[width=\linewidth]{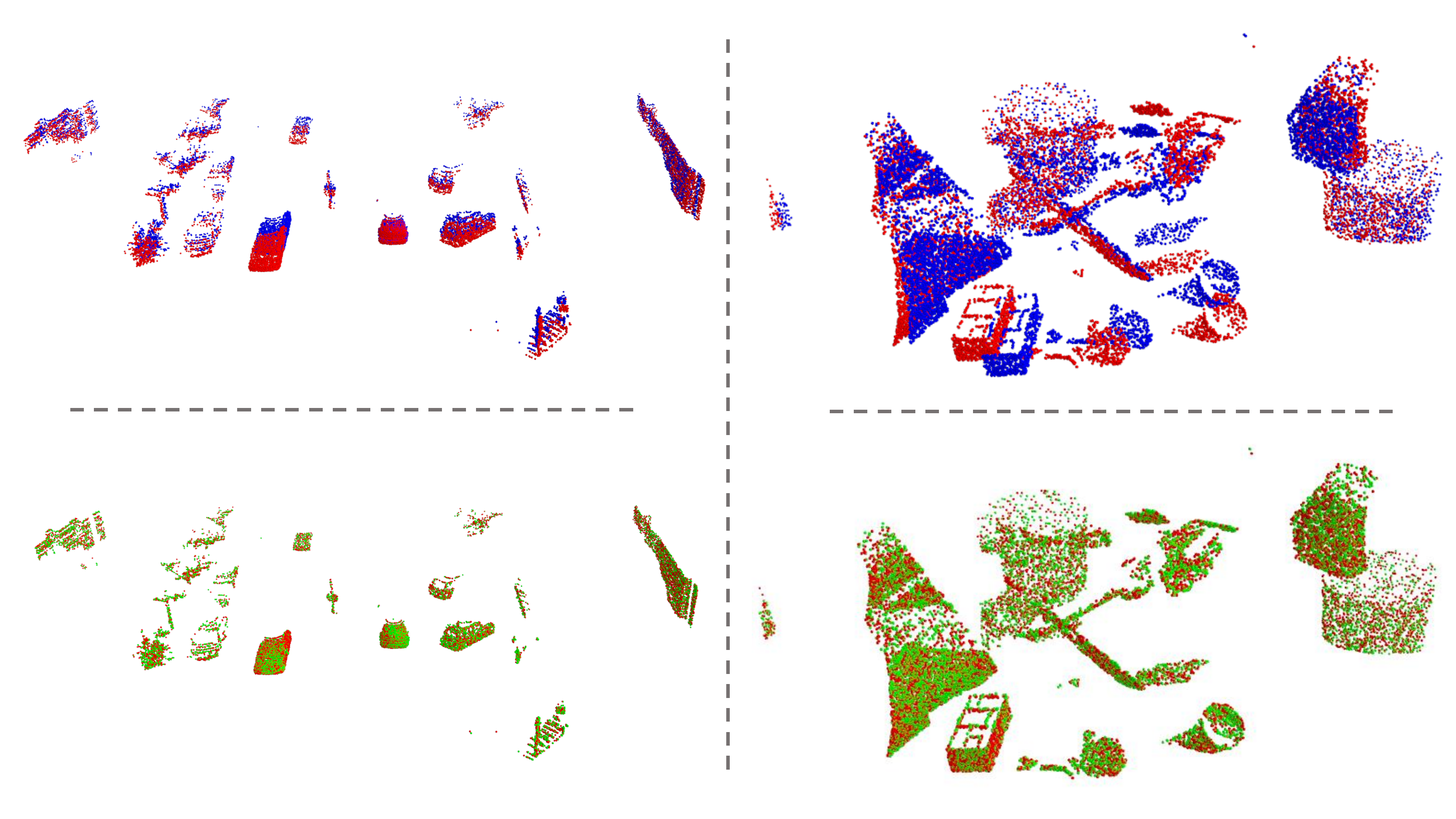} 
\caption{Qualitative results on the KITTI dataset(left) and FlyThings3D dataset (right) of non-occluded version. Top: source point cloud (blue) and target point cloud (red). Bottom: warped point cloud (green) utilizing the estimated flow of source point cloud and target point cloud (red).} 

\label{figs} 
\end{figure}

\begin{figure}[htb]
\centering 
\includegraphics[width=\textwidth]{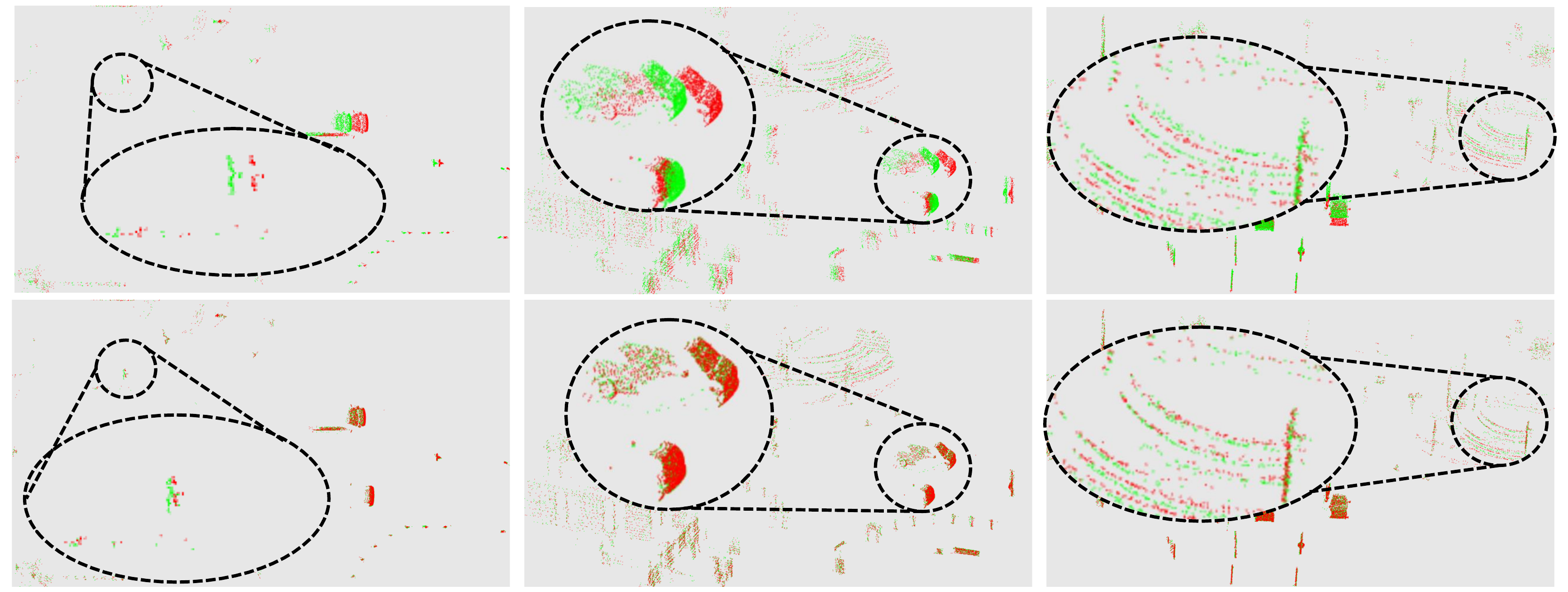} 
\caption{Qualitative results on the KITTI dataset of occluded version. Top: source point cloud (green) and target point cloud (red). Bottom: warped point cloud (green) utilizing the estimated flow of source point cloud and target point cloud (red).} 
\label{figo} 
\end{figure}

\begin{table}[htb]
\begin{center}
\caption{Performance comparison on the FlyingThings3Do and KITTIo datasets. All the models in the table are only trained on the occluded Flyingthings3D and tested on the occluded KITTI without any fine-tune. The best results for each dataset are marked in bold. }

\label{table:occ}  

\begin{tabular}{c|l|llll}
\hline
Dataset & Method & EPE(m)${\downarrow}$ & Acc Strict${\uparrow}$ & Acc Relax${\uparrow}$  & Outliers${\downarrow}$ \\
\hline

\multirow{7}{*}{FlyThings3Do} 
& FlowNet3D\cite{liu2019flownet3d} & 0.1577 & 0.2286 & 0.5821 & 0.8040 \\

& PointPWC-Net\cite{wu2020pointpwc}  & 0.1552 & 0.4160 & 0.6990 & 0.6389 \\

& SAFIT~\cite{shi2022safit}  & 0.1390 & 0.4000 & 0.6940 & 0.6470 \\

& OGSF\cite{ouyang2021ogsf}  & 0.1217 & 0.5518 & 0.7767 & 0.5180 \\


& 3D-OGFLow\cite{Ouyang2020OcclusionGS}  & 0.1031 & 0.6376 & 0.8240 & 0.4251 \\

& Estimation-Propagation\cite{Wang2022EstimationAP}  & 0.0781 & 0.7648 & 0.8927 & \textbf{0.2915} \\

& GMA3D & \textbf{0.0703} & \textbf{0.7908} & \textbf{0.9223} & 0.3101 \\
\hline

\multirow{7}{*}{KITTIo} 
& FlowNet3D\cite{liu2019flownet3d} & 0.1834 & 0.0980 & 0.3945 & 0.7993 \\

& PointPWC-Net\cite{wu2020pointpwc}  & 0.1180 & 0.4031 & 0.7573 & 0.4966 \\

& SAFIT~\cite{shi2022safit} & 0.0860 & 0.5440 & 0.8200 & 0.3930 \\

& OGSF\cite{ouyang2021ogsf}  & 0.0751 & 0.7060 & 0.8693 & 0.3277 \\


& 3D-OGFLow\cite{Ouyang2020OcclusionGS}  & 0.0595 & 0.7755 & 0.9069 & 0.2732 \\

& Estimation-Propagation\cite{Wang2022EstimationAP}  & 0.0458 & 0.8726 & 0.9455 & \textbf{0.1936} \\

& GMA3D & \textbf{0.0385} & \textbf{0.8997} & \textbf{0.9651} & 0.1986\\
\hline

\end{tabular}

\end{center}
\end{table}

\subsection{Performance on FT3D and KITTI without Occlusion}
We compared the results obtained for GMA3D on datasets FT3Ds and KITTIs with previous methods~\cite{gu2019hplflownet,wu2020pointpwc,puy2020flot,wei2021pv}, and detailed comparison results are shown in Table~\ref{table:s}. We increase the number of GRU iterations to 12 and epochs to 45 in training unlike the baseline~\cite{wei2021pv}, so the model can better integrate the original motion information and the motion information obtained by self-similarity aggregation. Experiments show that our GMA3D module achieves promising results and outperforms the baseline in terms of EPE by $13.9\%$ and $22.5\%$ on FT3Ds and KITTIs datasets respectively, which demonstrates that GMA3D can still produce more beneficial solutions to the non-occlusion scene while solving the occlusion problem, as revealed in Figure~\ref{figs}.

Through our experiments, we concluded that there are two reasons why GMA3D improves the performances on the non-occlusion version of the datasets: First, from the analysis of occlusion above, we infer that the farthest point sampling algorithm may cause the deletion of local matching areas between two consecutive point clouds, leading to the occurrence of hidden occlusion. Secondly, our GMA3D module aggregates local and global motion information through self-semantic-similarity, which not only can smooth local motion, but also decrease the motion inconsistency of local areas in the first point cloud.

\subsection{Performance on FT3D and KITTI with Occlusion}
We also compare our GMA3D module with the previous methods~\cite{besl1992method-icp,liu2019flownet3d,gu2019hplflownet,puy2020flot,wu2020pointpwc,ouyang2021ogsf} on the datasets FT3Do and KITTIo. We trained the GMA3D module on n = 8192 points with an initial learning rate of 0.001 for 45 epochs, 12 iterations, and the refine module for 10 epochs, 32 iterations. The detailed results of the comparison are shown in Table~\ref{table:occ}. In the synthetic FT3Do dataset, the performance of our GMA model is basically the same as that of the state-of-the-art~\cite{Wang2022EstimationAP} method. However, GMA3D has a stronger generalization ability, which performs well on the real dataset KITTIo without any fine-tuning. And~\cite{Wang2022EstimationAP}\cite{ouyang2021ogsf}\cite{Ouyang2020OcclusionGS} relies on the ground truth of the occluded mask, which is challenging to obtain in the real world. In contrast, our GMA3D only depends on the 3D coordinates of the point clouds and exhibits greater competitiveness in the real world. Figure~\ref{figo} visualizes the effect of GMA3D on scene flow estimation of occlusion points in the KITTIo dataset.






\begin{table}[htb]
\begin{center}
\caption{ Ablation Studies of GMA3D on the KITTIo dataset}
\label{table:ab}
\begin{tabular}{c|l}
\hline
Method                                                              &  EPE(m) $\downarrow$ \\
\hline
\hline
Backbone w/o GMA3D                                                   &  0.1084 \\
\hline
GMA3D w/o offset aggregator (original aggregator = MLP)              &  0.0412 \\

GMA3D w/o offset aggregator and Local Similarity Map                 &  0.0882 \\

GMA3D w/o offset aggregator and Global Similarity Map                & 0.0803 \\

GMA3D (full, with offset aggregator and Local-Global Similarity Map) &  \textbf{0.0385} \\
\hline
\end{tabular}

\end{center}
\end{table}

\subsection{Ablation studies}

We conducted experiments on FT3Do datasets to testify to the effectiveness of various modules in the GMA3D, including the offset aggregator and LGSM module. We gradually add these modules to GMA3D, and the final results are shown in Table \ref{table:ab}. From Table \ref{table:ab}, we can deduce that each module plays an essential role in GMA3D. First, The model does not perform well when the offset aggregator is not introduced. This is because the original transformer is designed for the domain of natural language processing. However, there are many differences between natural language and point clouds, so it is unable to be directly applied to point clouds. Secondly, we find that only focusing on the global motion information will produce poor results. With a local-global self-similarity map, GMA3D can improve accuracy by aggregating motion features from local and global aspects, respectively.

\section{Conclusion}
In this work, we proposed GMA3D to figure out the motion occlusion in scene flow from a local-global motion aggregation approach. GMA3D utilizes local and global self-attention mechanisms to aggregate motion features to infer the motion information of local and global occluded points in the first point cloud. In addition, GMA3D can smooth local motion, which is also beneficial to the scene flow estimation of non-occluded points. Experiments performed on both occluded and non-occluded datasets verify the superiority and generalization ability of our GMA3D module.

\bibliographystyle{splncs04}
\bibliography{ref}

\end{document}